\documentclass[journal,english]{IEEEtran}
\usepackage{tikz}
\usepackage[T1]{fontenc}
\usepackage{babel}
\usepackage{siunitx}
\usetikzlibrary{shapes,shadows,arrows,calc}
\usepackage{verbatim}
\usepackage{graphicx}
\usepackage{mathtools}
\usepackage{tkz-euclide}
\usepackage{cite}
\usepackage[T1]{fontenc}
\usepackage[utf8]{inputenc}
\usetkzobj{all}
\usepackage{subcaption}
\usepackage{url}
\usepackage{amssymb}

\title{Automatic Assessment of Artistic Quality of Photos}
\author{\IEEEauthorblockN{$Ashish ~Verma^1$, $Kranthi ~Koukuntla ^2$, $Rohit ~Varma ^3$ and $Snehasis ~Mukherjee^4$}

\IEEEauthorblockA{Indian Institute of Information Technology Sri City\\$ashish.v13@iiits.in^1$, $kranthi.k13@iiits.in^2$, $rohit.d13@iiits.in^3$, $snehasis.mukherjee@iiits.in^4$}}

\begin{document}
\IEEEoverridecommandlockouts
\maketitle
\textbf{}
\begin{abstract}
This paper proposes a technique to assess the aesthetic quality of photographs. The goal of the study is to predict whether a given photograph is captured by professional photographers, or by common people, based on a measurement of artistic quality of the photograph. We propose a Multi-Layer-Perceptron based system to analyze some low, mid and high level image features and find their effectiveness to measure artistic quality of the image and produce a measurement of the artistic quality of the image on a scale of 10. We validate the proposed system on a large dataset, containing images downloaded from the internet. The dataset contains some images captured by professional photographers and the rest of the images captured by common people. The proposed measurement of artistic quality of images provides higher value of photo quality for the images captured by professional photographers, compared to the values provided for the other images.
\end{abstract}
\IEEEoverridecommandlockouts
\begin{IEEEkeywords}
Multi-Layer-Perceptron, Image Artistic Quality
\end{IEEEkeywords}

\section{Introduction}
\noindent Classifying images based on visual asthetics between two classes, viz, image shot by professional and image shot by non-professional has been a trend in the recent past \cite{tang}. The recent advancements in technology and popularization of digital camera, have resulted in producing a big pool of unassessed photographs. Algorithms which can assess and grade the image, can be useful in ranking the image, and this image ranking algorithm can enhance the result of image based search engine, which can be useful in various tasks. For example, the newspaper editor may ask for a good quality photograph from a large pool of images available in the internet, regarding an incident. Also, with the growing trend of people to upload photographs in social media websites, gives rise to the necessity of an algorithm capable of finding the best quality photograph related to a specific theme.

Various parameters are found to be effective for assessing visual aesthetics to grade the artistic quality of the image \cite{tang}. In this study, we have experimentally identified such parameters and introduced a multi-layer perceptron (MLP) based technique with the features extracted with the parameters, to come up with a score that measures the artistic quality of the query image. Following are the parameters used for defining asthetics of an image.

\noindent \textbf{Composition:} Photographic composition is the arrangement of visual elements in a photo. Photographers consider a large variety of elements like lines, shape of objects, patterns, texture etc. In artistic photography, simplicity in color composition in the photo is often encouraged \cite{art}. Moreover, the color arrangement and contrast along diagonal lines, curved lines in the image have effects on the artistic quality. Also, the locations of the objects in the image, influence the photographic quality.

\noindent \textbf{Lighting:} Lighting condition of a photograph plays a vital role in its artistic quality. The amount of light and direction of light source are important factors in measuring artistic quality of photos. Weather plays an important role in natural photography, whereas in the studio settings, direction of light source becomes an important cue for making stunning effects, as well as, enhancing 3D impression of objects using shadows. The contrast of the object also helps in emphasizing the area of interest in the photo.

\noindent \textbf{Color Arrangement:} Professional photographers use various techniques to control colors and use specific combinations of colors influencing specific emotions of the viewers \cite{1,2}. Light colors, for example blue, yellow tend to create a soothing effect, whereas cold colors like gray, red tend to create a scary 
effect or calming effect on the viewers. Photographers take leverage of this effects on emotions to make a photograph more expressive about the story behind the photo.

\noindent \textbf{Blur}: In a photograph taken by professional photograph, blur comes rarely. It actually happens because of glitches like poor equipment or a vibration in the camera or because of the quality of lens.

\noindent \textbf{Contrast:} Low contrast leads to poor quality of photograph. Amateur photographers tend to use poor equipment due to which they often come up with low contrast photos.

\begin{figure*}
\centering{
\includegraphics[height=3.3cm, width=4.2cm]{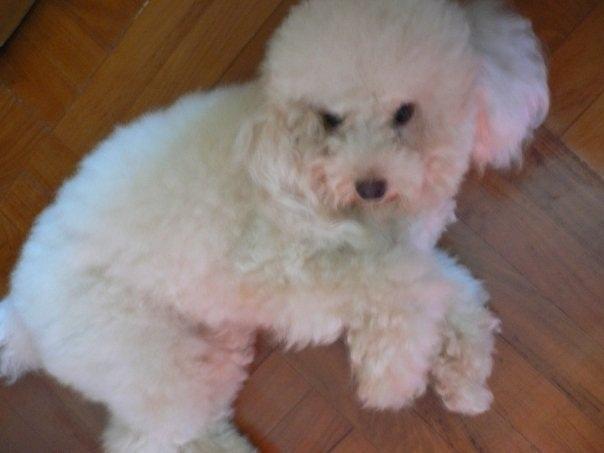}
\includegraphics[height=3.3cm, width=4.2cm]{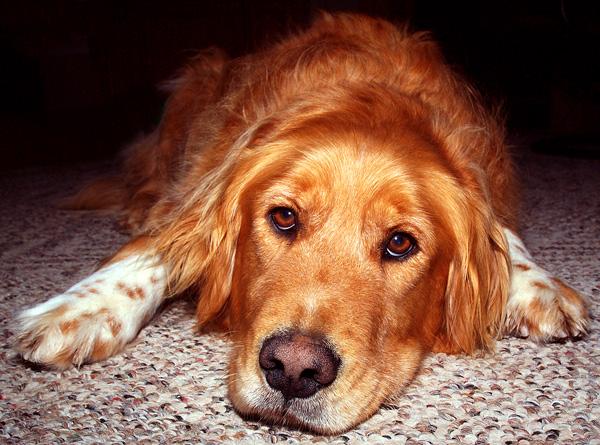}
\includegraphics[height=3.3cm, width=4.2cm]{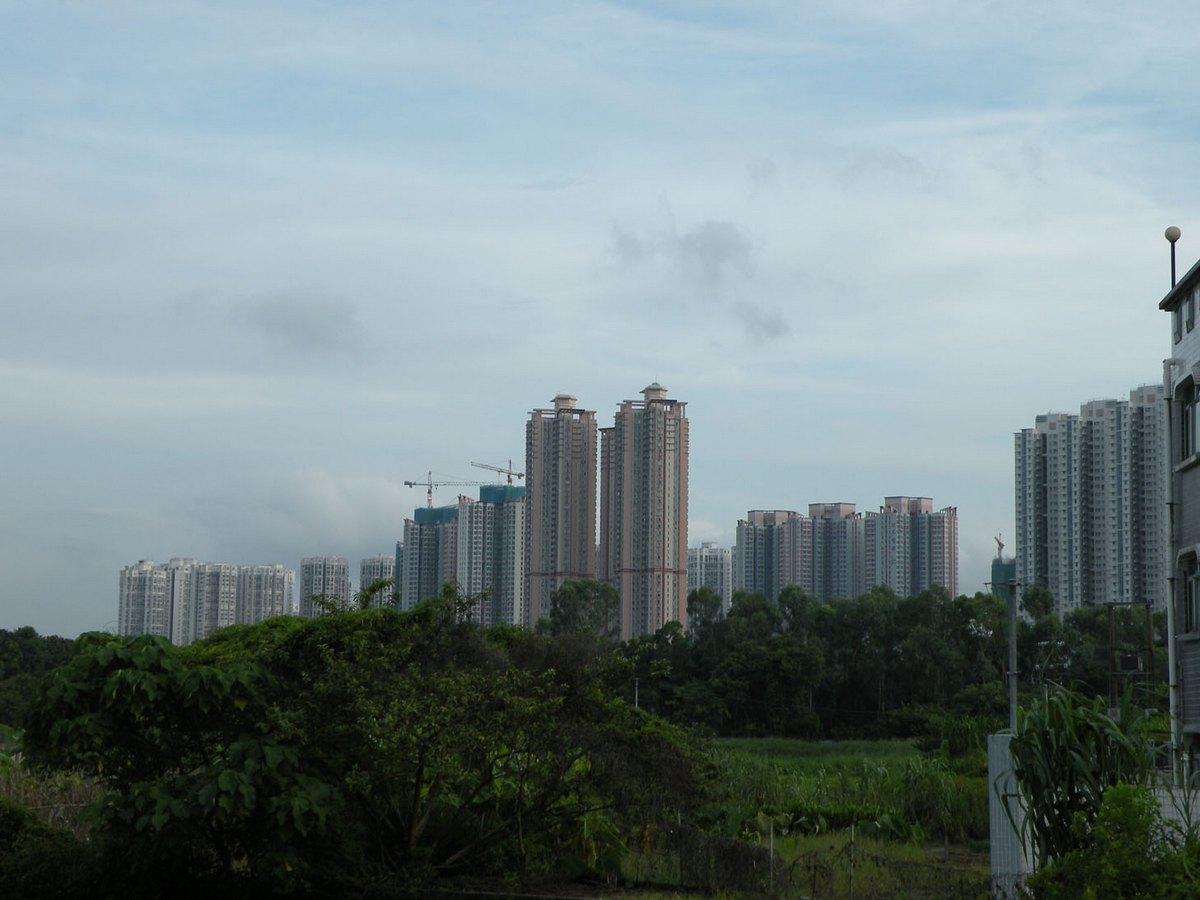}
\includegraphics[height=3.3cm, width=4.2cm]{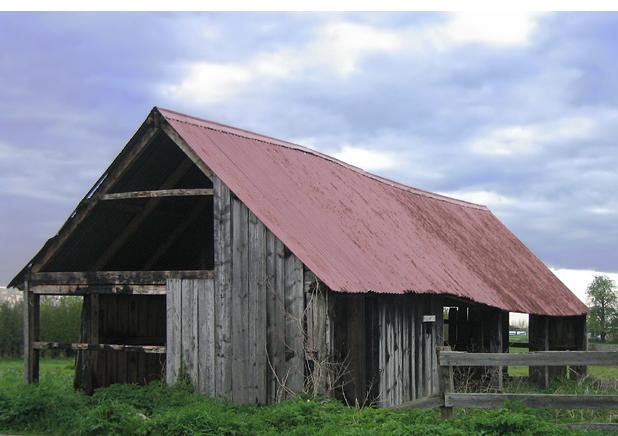}}
\centerline{(a) score:3.23\hspace*{0.75in}(b) score:7.4\hspace*{0.75in}(c) score:3.72\hspace*{0.75in}(d) score:6.52}
\centering{
\includegraphics[height=3.3cm, width=4.2cm]{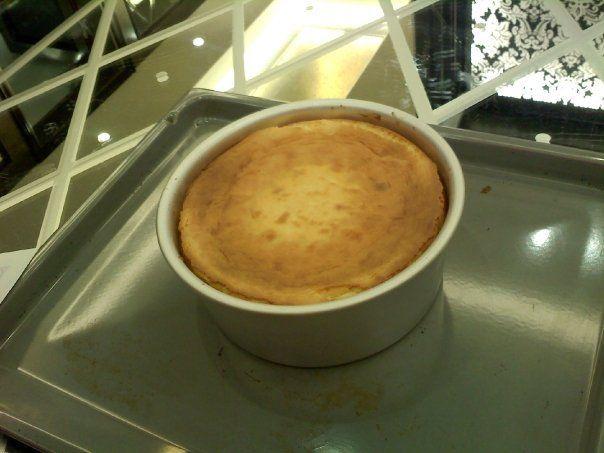}
\includegraphics[height=3.3cm, width=4.2cm]{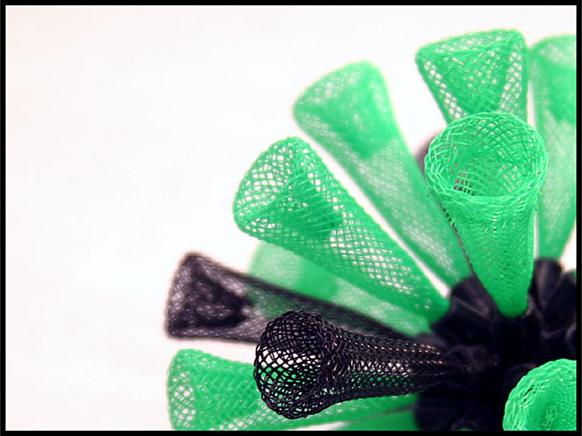} \includegraphics[height=3.3cm, width=4.2cm]{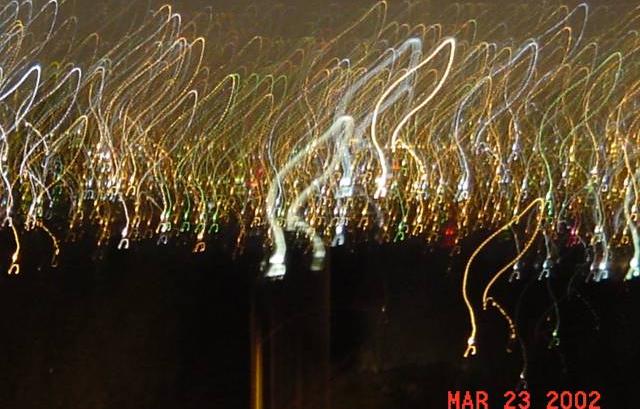}
\includegraphics[height=3.3cm, width=4.2cm]{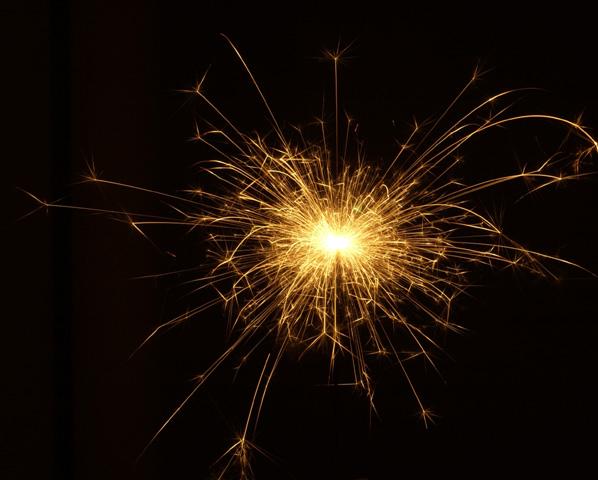}
}
\centerline{(e) score:1.44\hspace*{0.75in}(f) score:8.34\hspace*{0.75in}(g) score:1.63\hspace*{0.75in}(h) score:4.07}
\centering{
\includegraphics[height=3.3cm, width=4.2cm]{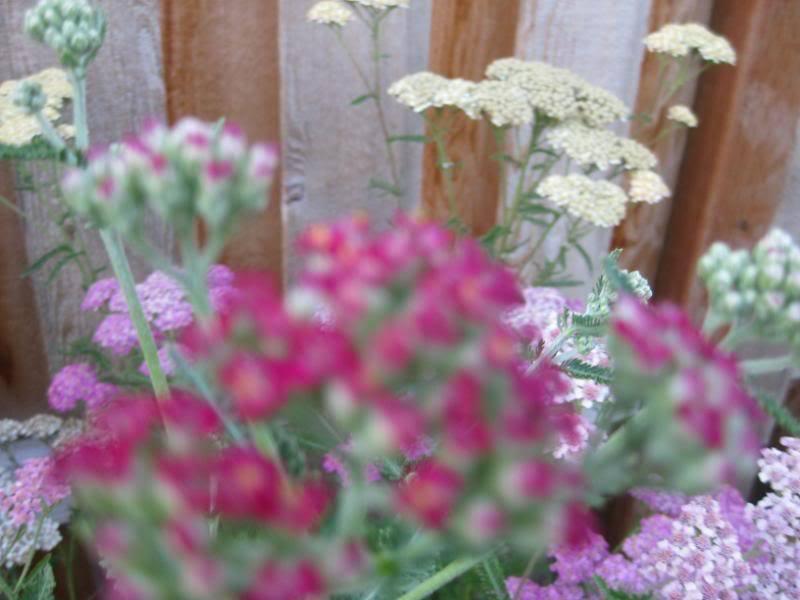}
\includegraphics[height=3.3cm, width=4.2cm]{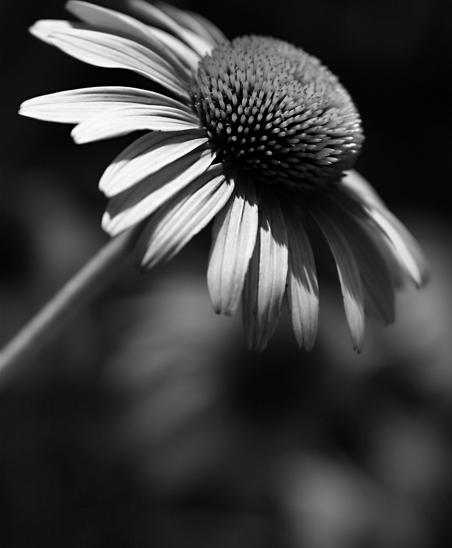}
\includegraphics[height=3.3cm, width=4.2cm]{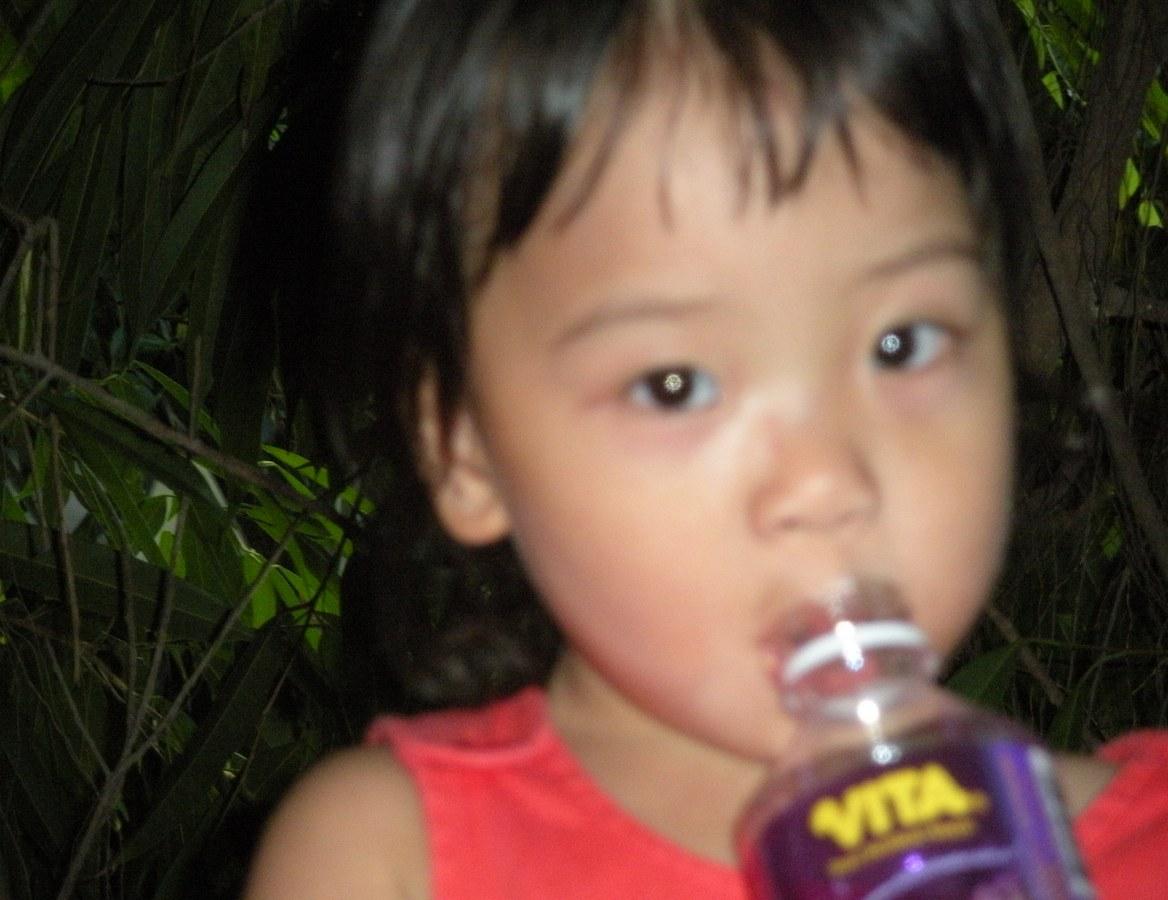}
\includegraphics[height=3.3cm, width=4.2cm]{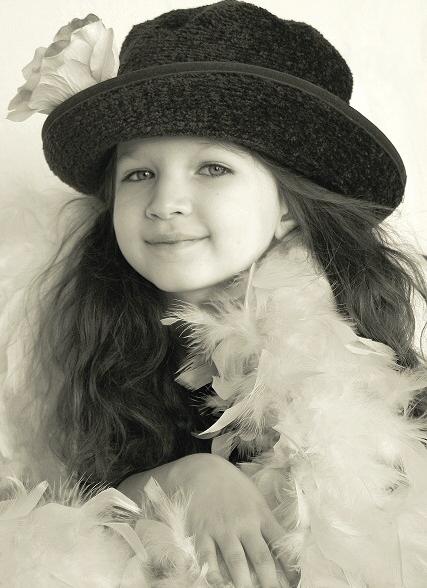}
}
\centerline{(i) score:2.32\hspace*{0.75in}(j) score:8.87\hspace*{0.75in}(k) score:3.61\hspace*{0.75in}(l) score:7.34}
\centering{
\includegraphics[height=3.3cm, width=4.2cm]{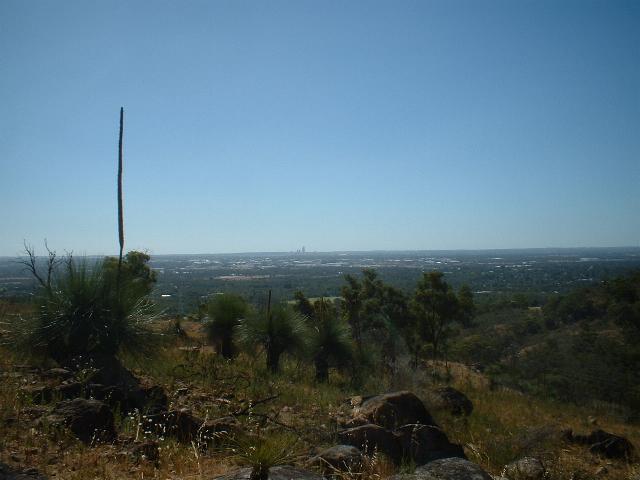}
\includegraphics[height=3.3cm, width=4.2cm]{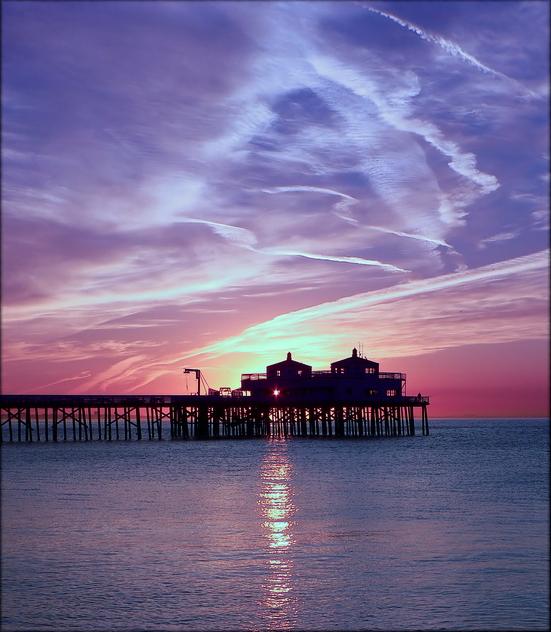}
}
\centerline{(m) score:2.95\hspace*{0.75in}(n) score:4.76}
\caption{Example photographs taken from \cite{tang} dataset, alongwith the quality scores provided by the proposed method. The categories of photographs are given as follows: (a,b) Animal, (c,d) Architecture, (e,f) Static, (g,h) Night, (i,j) Flower, (k,l) Person, (m,n) Landscape. For each category, first one represents a photograph captured by common people and the second one represents photograph taken by professional photographer.}
\label{exp}
\end{figure*}

The main contribution of this paper is to introduce a MLP based model which takes all the above parameters of artistic photography into consideration, provides suitable weightage to each parameter to come up with a measure of artistic quality of the image. We applied the proposed method on a large database of annotated photographs taken by both professionals and common people \cite{tang}. The dataset consists of seven categories of photographs, almost equaly distributed as photographs captured by professionals and common people. Figure \ref{exp} shows examples of images taken from the dataset introduced in \cite{tang}, alongwith score provided by the proposed method. Clearly, the photographs captured by professionals secured higher value according to the proposed method, compared to the photographs captured by the common people.

The rest of this paper is organized as follows. Section \ref{feature} discusses the process of extracting suitable features from the images, corresponding to the above parameters. Section \ref{mlp} describes the proposed MLP based model to provide an effective feature combining all the parameters. The results of applying the proposed method on a large dataset is discussed in Section \ref{result}, followed by conclusions and future scopes of research in Section \ref{conclude}. Before all these we briefly discuss the state-of-the-art in the area of photo quality assessment.

\section{Related Works}
Several efforts have been made in the recent past on assessment of image quality \cite{cvpr15}, based on semantics. However, two class classification of photos, based on the artistic quality, is a less-explored area of research. Some of the existing methods classify the photographs and graphics \cite{3,7}, whereas some are able to classify indoor and outdoor photographs \cite{8,11}. Efforts have been made in classifying photographs and paintings \cite{6,9}. In most of the above classification problems, image intensity based features play a vital role in classification. However, classification of photographs taken by professional photographers or common people, is a more challenging and less-studied problem, as low level intensity based features cannot differentiate between the two classes. In order to incorporate the parameters discussed in the introduction, a suitable combination of low, mid and high level image features are needed.

Luo et. al. \cite{14} proposed a method for content-based photo quality assessment, where local features extracted from the interest regions of the images, are combined with the global features extrated from the whole image, to find the object-background correlation. However, some important features like blurness, brightness, etc. are ignored in the study. Zhang et. al. proposed another approach combining global and local image features to predict structural cues of the objects, for quality assessment \cite{23}. Lo et. al. \cite{icpr12} combined some hand crafted features extracted from the images like, dark chanel prior, edge sharpness, color distribution, etc. Position and scale of the object of interest in a photograph, is an important parameter to assess the photograph, which is ignored in \cite{icpr12}. Bhattacharya et.al. proposed more sophisticated features for photo quality assessment, considering the position and scale of the object of interest, by calculating the distance of the object from the four spatial points of the image \cite{mm10}. The four spatial points determine the center location of the photo. However, the clarity of the photograph is not emphasized in \cite{mm10}.

Efforts have been made to extract several low, mid and high level image features, combine them and use SVM to test the effectiveness of the features, for photo quality assessment \cite{datta, dhar}. Datta et. al. \cite{datta} extracted 56 features based on human intuition and ran SVM classifier. The photo aesthetics are predicted using classification tree and linear regression. Dhar et. al. \cite{dhar} combined multiple features related to the aesthetics and built a system for predicting attractiveness of the photo based on attributes such as layout of objects in the image, contents of the image and illumination. The intution based features often ignore the low level semantics of the photographs, which are useful for the assessment of quality. To overcome the limitations of intuition-based features, some low level features like GIST, SIFT, HOG, etc. have been used to measure the photo aesthetics \cite{20, 21}. Bag of Features (BoF) technique is applied on the low level features to assess the quality of photo in \cite{20}. In \cite{21}, the low level features are fed in to a multiclass classifier for assessing the quality. However, the handcrafted features used in \cite{20, 21} do not cover all the parameters for photo aesthetics, as given in the Introduction. Cao et. al. proposed an adaptive learning technique to rank the photographs based on visual aesthetics and applied on face images \cite{sp}. However, for scene images such features cannot be applicable. Efforts have been made to enhance the photo quality using seam carving technique \cite{sp1}, but ranking photographs based on aesthetic quality is still an unvisited area of research.

An efficient set of image features, combining the concepts of clarity, color combination and position and scale of the object of interest, has been proposed by Ke et. al. \cite{15} and Tong et. al. \cite{13}. Tang et. al. combined the features of \cite{13}, \cite{14} and \cite{15} and applied SVM classifier to classify the images into two categories: photographs taken by professional photographer and by common people \cite{tang}. All the parameters considered for artistic evaluation of photographs are covered by the features used by \cite{tang}. However, the main disadvantage in \cite{tang, 13, 14, 15} is that, all the features corresponding to the parameters for photo quality assessment, as discussed in the introduction, are assumed to have equal contributions to the assessment of photo. However, in reality, professional photographers suggest different priorities for the parameters. For example, image clarity should be given more priority than the light effect in the photograph \cite{art}. In this study, we have addressed this limitation by using the useful image features in a Multi Layer Perceptron (MLP) based framework. We train the proposed MLP based system for suitable weightage to each of the image features. Moreover, we provide a measure of quality, based on the proposed MLP based system, to rank the photos in terms of the artistic quality. Next we describe the process of feature extraction.

\section{Feature Extraction}
\label{feature}
We extract features from the photographs, to emphasize on the parameters responsible for maintening the quality of a photograph, as discussed in the introduction. These features can be helpful for bridging the gap between human perception and machine computation.

\subsection{Spatial Distribution of Edges} 
Simplicity in a photograph can be measured by computing the spatial distribution of the high frequency edges of an image. The background of low quality photos or snapshots are often cluttered. High frequency edges for the objects of interest are often found in professional photos since the subject of the photograph should be focused and well defined. Hence, for professional photographs, the objects of interest are expected to be around the centre of the image and the specified location around the object should have high frequency.

In order to measure such criteria to be a good photograph, we apply a $3\times 3$ Laplacian filter with $\alpha = 0.2$ on the image, and take its absolute value to ignore the direction of the gradients. For color images, mean is found across the channels after filtering each of the red, green and blue channels that are applied separately. Finally, we resize the Laplacian image size to $100\times 100$ and normalize the image sum to 1. The advantage of resizing Laplacian image size is to calculate the edge distribution in images and study the differences between high quality photos and low quality snapshots. Let $M_{p}$ and $M_{s}$ be the mean Laplacian images of the professional photos and snapshots, respectively. We use $L_{1}$ distance to measure the distance between the probe’s Laplacian image, $L$ and the mean Laplacian images.

The quality of the probe image is defined as:
\begin{equation}
q_{l} = d_{s} - d_{p},
\end{equation}
where $d_s$ and $d_p$ are given by,
\begin{equation}
d_{s} = \sum \limits_{x,y}|L(x,y) - M_{s}(x,y)|, 
\end{equation}
\begin{equation}
d_{p} = \sum \limits_{x,y}|L(x,y) - M_{p}(x,y)|.
\end{equation}
Measuring the area occupied by the object of interest, is a good way to measure the compactness of the spatial distribution of edges. The area of the bounding box that encloses the top $\delta$ percentage of the edge energy is calculated, where the value of $\delta$ is set to 96.04, as in \cite{tang}.

\subsection{Color Distribution}
Professional photographers often use color palette, which is an important parameter to identify the differences between photographs taken by professional photographers and non-photographers. We obtain a histogram based on the color distributions of the given photograph. The histogram is obtained by quantizing the span of intecsity values in red, blue and green channels separately, into 16 bins each. Hence, each photograph is represented by a 4096 = $16^{3}$ bin histogram. The histogram is normalized to unit length. We apply kNN classifier to distinguish whether the probe image is more like a professional photo or a snapshot. We use L1 metric to calculate the distance between histograms, which is found to give the best result. We keep k = 5 in our study, as done in \cite{15}. We calculate the quality score $q_{cd}$ for the photograph, with respect to color distribution as follows:
\begin{equation}
q_{cd} = n_{p} - n_{s},
\end{equation}
where $n_{p}$ and $n_{s}$ are the number of neighbors that are closer to professional photos and snapshots, respectively.

\subsection{Hue Count}
The hue count of a photograph determines its simplicity \cite{14}. Most good quality photographs look more brilliant than snapshots, which is depicted by the hue count of the photographs. To measure the quality of photograph according to hue count, color images are converted to its HSV representation and pixels with brightness values in the range [0.15,0.95] are taken into consideration and saturation $s > 0.2$. A 20-bin histogram $H$ is computed on the good hue values.

Let $m$ be the maximum (peak) value of the histogram and $N$ be the set of bins with values greater than $\alpha m$. Then,
\begin{equation}
N = \{i|H(i) > \alpha m\}.
\end{equation}
The quality of a photo is $q_{h}$ is calculated as,
\begin{equation}
q_{h} = 20 - ||N||.
\end{equation}
$\alpha$ controls the noise sensitivity of the hue count and we set
$\alpha = 0.05$ experimentally.

\subsection{Blur}
A good photograph should not be blurry at the region of interest \cite{art}. Hence, blur estimation at the region of interest of a photograph, is important for measuring its quality. Several methods have been proposed in the literature, for estimating blur of an image \cite{10,12}. We apply Tong et. al.’s blur estimation technique \cite{12} in our study, for its simplicity. We extract the region of interest of the photograph and find the blur as follows:

We model a blurred image $I_{b}$ as the result of a Gaussian smoothing filter $G_{\sigma}$ applied to a sharp image $I_{s}$, i.e.,
\begin{equation}
I_{b} = G_{\sigma} * I_{s}.
\end{equation}
We would like to recover the smoothing parameter $\sigma$ given
only the blurred image $I_{b}$. The image quality would be inversely proportional to $\sigma$. Let us assume that the frequency
distribution for all sharp photos $I_{s}$ is approximately the
same. We can estimate the maximum frequency of the image $I_{b}$ by taking its two dimensional Fourier transform and counting the number of frequencies whose power is greater than some threshold $\theta$. In other words, let the two dimensional Fourier transform be denoted by
\begin{equation}
F = FFT(I_{b}).
\end{equation}
Let the set of frequencies present in $I_{b}$ be denoted by
\begin{equation}
C = {(u,v)||F(u,v) > \theta}.
\end{equation}
Since the Gaussian filter is a low pass filter, the maximum frequency present in the image is equal to $||C||$. Thus, we define the image quality as follows:
\begin{equation}
q_{f} = \dfrac{||C||}{||I_{b}||} \sim \dfrac{1}{\sigma}.
\end{equation}
The parameter $\theta$ is introduced because a Gaussian filter does not produce a sharp cutoff of high frequencies. We experimentally set $\theta = 5$.

\subsection{Low Level Features}
For photo quality assessment low level features like contrast and brightness are important. Professional photos usually have higher contrast than snapshots. First, we compute the gray level histogram $E_{r}$, $E_{g}$ and $E_{b}$ for the red, green and blue regions respectively. Then, we compute the combined histogram $E$, where
\begin{equation}
E(i) = E_{r}(i) + E_{b}(i) + E_{g}(i).
\end{equation}
Professionals suggest that the exposure (brightness level) required for the subject and the background should be different \cite{art}. Therefore, the average brightness for the entire photo to deviate from 50 percentage gray by adjusting the exposure to be correct on the subject only. If the background is pure black or white, the deviation could be quite severe. The larger the deviation, the more likely that the photo was taken by a professional. We calculate the photo’s average brightness, $b$. If $b_i$ and $b_b$ are the average brightness at the object of interest and background respectively, then quality measure with respect to brightness of the photo is measured as:
\begin{equation}
b = b_i - b_b.
\end{equation}
Next we calculate the dark chanel prior.

\subsection{Dark Channel Feature}
The dark channel prior can be considered as an estimation of clarity, saturation, and hue composition. The dark channel estimation of a picture increments with how much it is obscured. A low-quality photo with dull shading gives a higher arrived at the midpoint of dull channel esteem. Moreover, extraordinary shade values provide distinctive dim channel values. So the dark channel feature additionally consolidates hue composition information.
\begin{figure*}
\centering{
\includegraphics[width=7in]{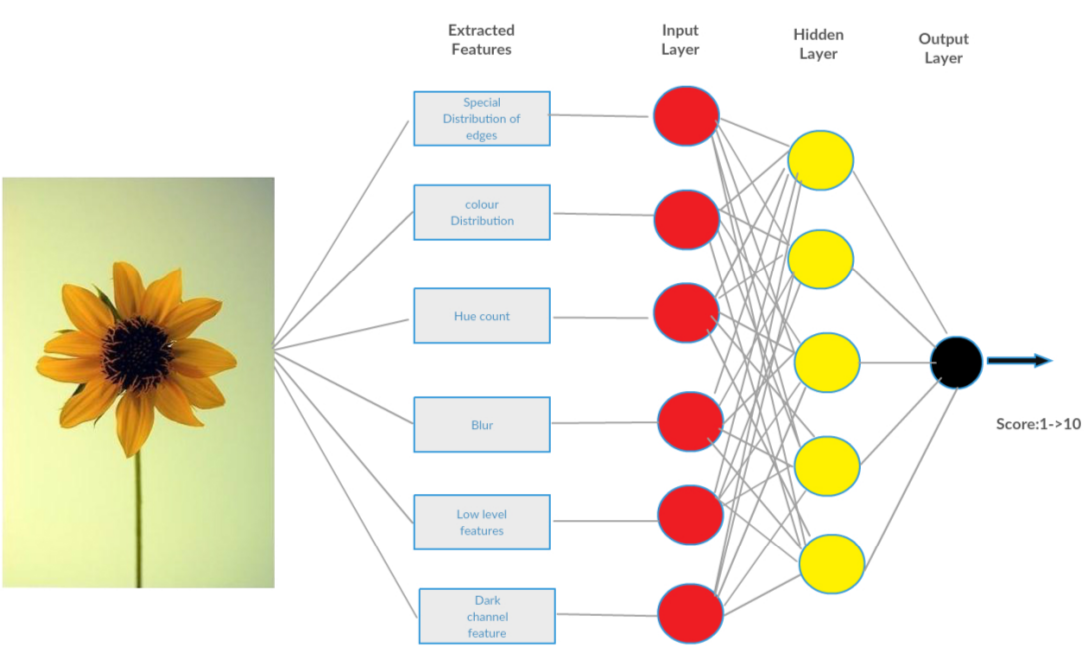}}
\caption{Overall structure of the proposed MLP based method.} 
\label{dgrm}
\end{figure*}

Dark channel prior was introduced by He et. al. \cite{4,5} for haze removal in images. The dark channel of an image $Im$ is defined as 
\begin{equation}
Im_{dark}(i) = min_{c\in R,G,B}(min_{i'\in \Omega(i)}I_{c}(i')) 
\end{equation}
where $Ic$ is a color channel of $I_{c}$ and $\Omega(i)$ is the neighbourhood of pixel i. We choose $\Omega(i)$ as a 10 * 10 local patch. To reduce the effect of brightness, we normalize the dark channel value by the sum of RGB channels. The dark channel feature of a photo $Im$ is computed as the average of the normalized dark channel values in the subject areas as follows:
\begin{equation}
\dfrac{1}{||S||} \sum \limits_{(i) \epsilon S} \dfrac{I_{dark}(i)}{\sum \limits_{c \epsilon R,G,B} I_{c}(i)},
\end{equation}
where $S$ is the subject area of $Im$. Next we discuss the proposed MLP based model.

\section{Proposed MLP Based Model}
\label{mlp}
We apply a Multi Layer Perceptron (MLP) based supervised learning model to combine all the features extracted following Section \ref{feature}, with proper weightage, to calculate a score for the photograph, as a measure of its artistic quality. We use a sigmoid activation function for every node, as follows:
\begin{equation}
\label{eqn}
f(t)=\dfrac{1}{1+e^{-t}}.
\end{equation}
The activation function $f(t)$ provides the quality score for the photograph. Here, $t$ is the weighted sum of all the features extracted from the photograph, as discussed in Section \ref{feature}.
\begin{equation}
t=\sum_i{q_iW_i},
\end{equation}
where $q_i$ is the $i$th feature value and $W_i$ is the corresponding weight. We introduce 6 neurons at the input layer, corresponding to the 6 features extracted from the image. We keep 5 neurons in the hidden layer. Figure \ref{dgrm} describes the overall MLP based procedure. We describe the training and testing procedures for the proposed system sequencially.

\subsection{Training}
For training, we start with equal weights for all the extracted features, and then update the weights in each iteration based on the feedback score. If the desired output is $o$ and the obtained output is $z$, then we define the objective function for the MLP model as follows:
\begin{equation}
J = \dfrac{1}{2}\sum_{k=1}^{c}(o-z)^2,
\end{equation}
where $c$ is the number of classes (here $c=2$).

We start with an equal edge weight for all the features and use gradient descent for each edge weight in each epoch (of the 250 epochs performed) to get a weight for which the objective function is minimum. If $W_{m}$ is the weight of an edge during $(m-1)^{th}$ epoch then $W_{m+1}$ which is the updated weight after $m_{th}$ epoch is given by,
\begin{equation}
W_{m+1} = W_{m}-\eta\dfrac{\delta J}{\delta W},
\end{equation}
where the value of the learning rate $\eta$ is set as 0.1.

Let $W_{kj}$ be the weight between $j^{th}$ hidden node and $k^{th}$ output node. We need to find $\dfrac{\delta J}{\delta W_{kj}}$ as the update on the edge weight between the hidden and the output layers of the MLP structure. Let $W_{ji}$ be the weight between $i^{th}$ input node and $j^{th}$ hidden node. Then the update of edge weight between the input and the hidden layer is given by, $\dfrac{\delta J}{\delta W_{ji}}$.

\subsection{Testing}
For test photographs, we calculate all the features described in Section \ref{feature}, and calculate the quality score using equation (\ref{eqn}). We normalize the quality score in a scale of 1 to 10 to produce the final quality score to the test photograph. We set a threshold value $\gamma$ on the quality score, to classify the photograph as a photograph taken by professional photographer or a common people. Next we discuss the experiments and results of applying the proposed method on a benchmark dataset.
\begin{figure*}
\centering{
\includegraphics[height=3.1cm, width=8.5cm]{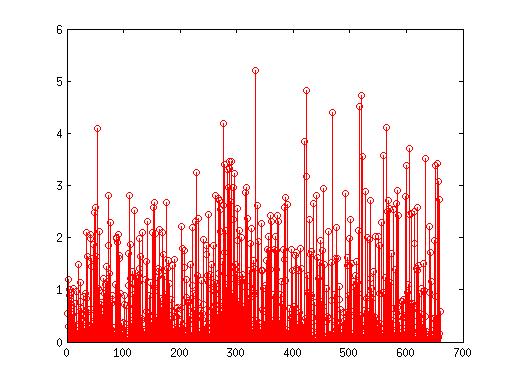}
\includegraphics[height=3.1cm, width=8.5cm]{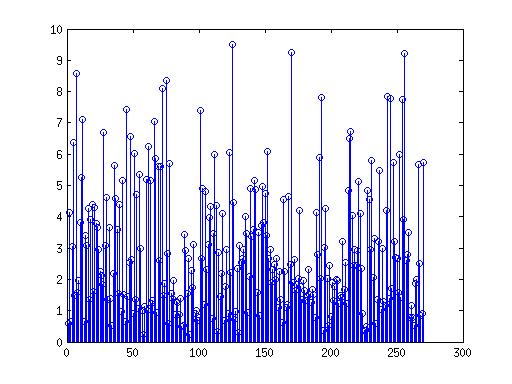}
\includegraphics[height=3.1cm, width=8.5cm]{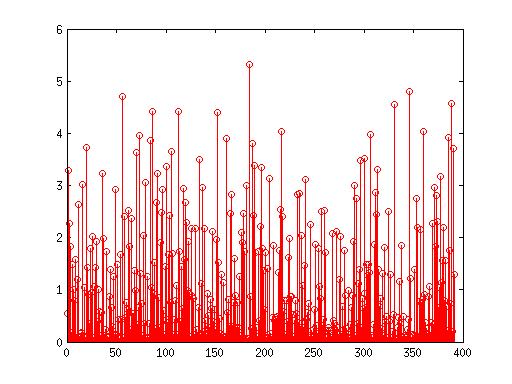}
\includegraphics[height=3.1cm, width=8.5cm]{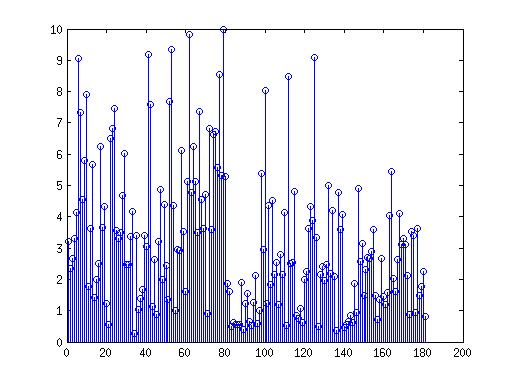}
\includegraphics[height=3.1cm, width=8.5cm]{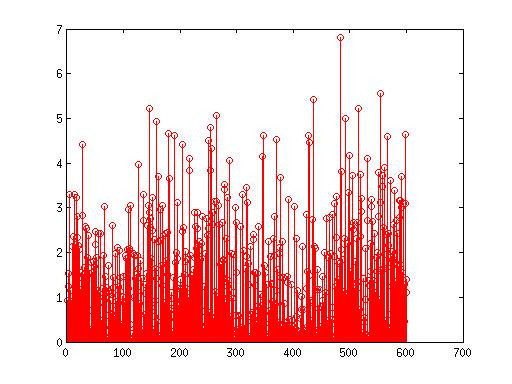}
\includegraphics[height=3.1cm, width=8.5cm]{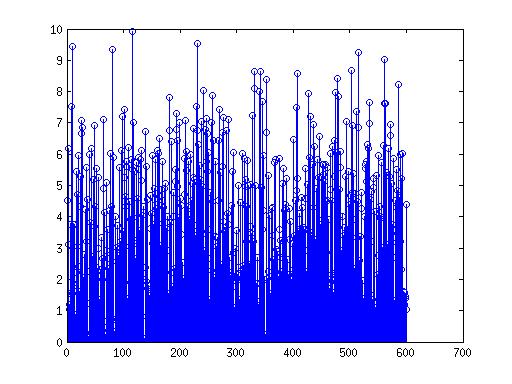}
\includegraphics[height=3.1cm, width=8.5cm]{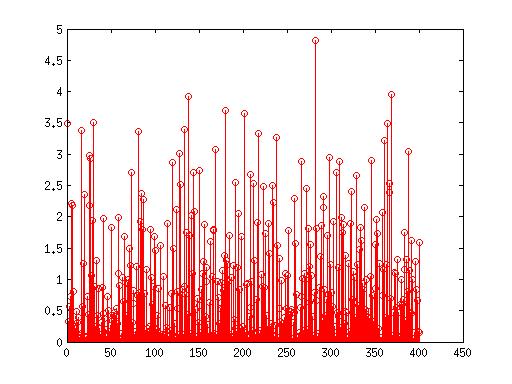}
\includegraphics[height=3.1cm, width=8.5cm]{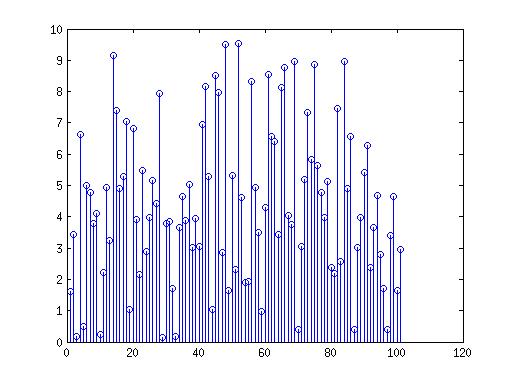}
\includegraphics[height=3.1cm, width=8.5cm]{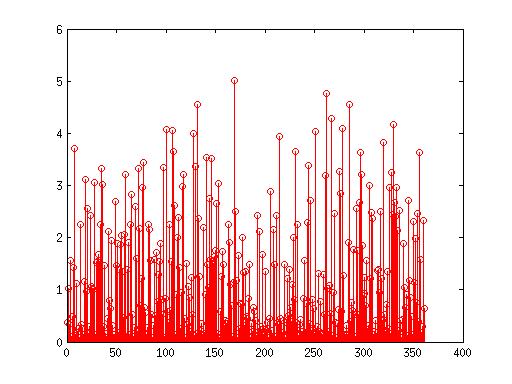}
\includegraphics[height=3.1cm, width=8.5cm]{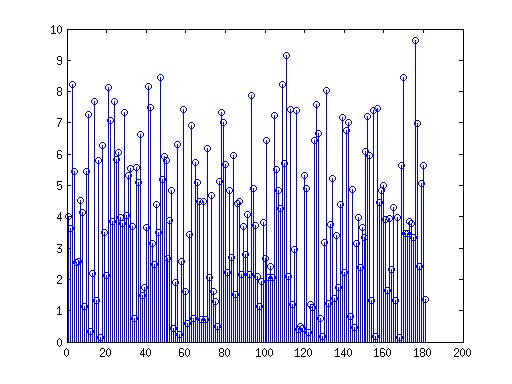}
\includegraphics[height=3.1cm, width=8.5cm]{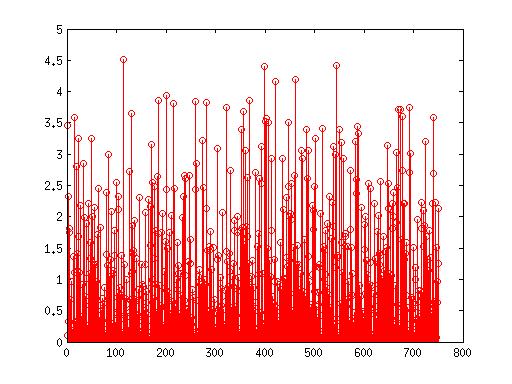}
\includegraphics[height=3.1cm, width=8.5cm]{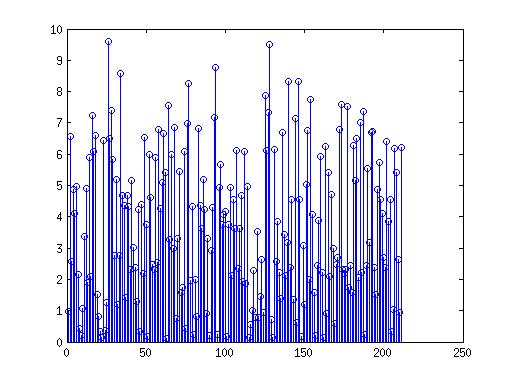}
\includegraphics[height=3.1cm, width=8.5cm]{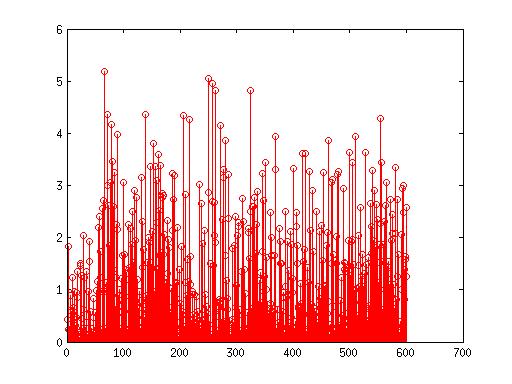}
\includegraphics[height=3.1cm, width=8.5cm]{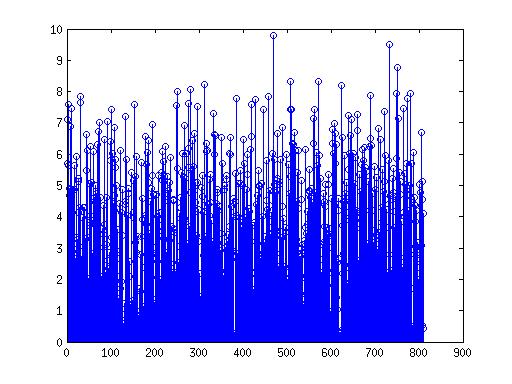}}
\centerline{(a)\hspace*{3in}(b)}
\caption{Quality scores obtained by all the test photographs taken from the dataset introduced by \cite{tang}. The categories of photographs are given as follows: row1 Animal, row2 Architecture, row3 Static, row4 Night, row5 Flower, row6 Person, row7 Landscape. For each category, (a) represents scores obtained by photographs captured by common people and (b) represents scores obtained by photographs taken by professional photographers. For each graph, x-axis represents the images and y-axis represents their corresponding quality scores as provided by the proposed method.}
\label{score}
\end{figure*}

\section{Experiments and Results}
\label{result}
In this section we discuss the experiments carried out to validate the proposed method, alongwith discussions on the results obtained. We describe the dataset used, followed by a description of the experimental set up. Finally we discuss and analyze the results.

\subsection{Dataset} 
For validation of the proposed method for assessment of photo quality, we found only one publicly available dataset, which is large enough and and diverse towards the contents of the photographs. The dataset is introduced by Tang et. al. \cite{tang}. The dataset includes 17673 images broadly classified into 7 different categories, viz., animal, architecture, human, landscape, night, plant and static, based on the visual content of the photographs. Each of the categories of photographs contains significant number of photographs of both the classes - photographs taken by professional photographers and by common people. Table I shows the distribution of 17673 images across various categories. We randomly choose approximately 70 percent of the images from each category for training and rest 30 percent we keep for testing purposes.
\begin{table}
\begin{center}
\caption{Distribution of 17673 images of the dataset across different categories.}
\renewcommand{\arraystretch}{1.05}
\begin{tabular}{|m{2cm}|m{2.5cm}|m{2cm}|}
\hline
\textbf{Category} & Taken by Professional photographers & Taken by common people\\ \hline
Animal & 953 & 2292\\ \hline
Architecture & 595 & 1290\\ \hline
Human & 678 & 2470\\ \hline
Landscape & 820 & 1950\\ \hline
Night & 353 & 1356\\ \hline
Plant & 594 & 1803\\ \hline
Static & 531 &2005\\ \hline
\end{tabular}
\end{center}
\label{data}
\end{table}

\subsection{Experiments} 
Extracted features were normalized before providing them as input to the MLP to provide equal weightage to each of the mentioned feature. We update the weights of each feature in each iteration based on the error. For training purposes, we assume that the photographs captured by professionals are of high quality and the others are of low artistic quality, and label the photographs accordingly, as done in \cite{tang}. We validate the proposed method on the 30\% of the images of each of the seven categories, kept for testing. We classify the photographs into two categories: high quality and low quality photographs based on the quality score given by the proposed method. If the quality score is greater than a threshold $\theta$, then we label the test image as a high quality image, otherwise it is a low quality image. We move the threshold value $\theta$ from 2 to 9 with 0.5 difference and for each case, the accuracy of the proposed method is measured. The proposed method gives the highest accuracy for $\theta=4$. Figure \ref{score} compares the quality scores obtained by the high quality and low quality photographs of each category, as provided by the test images. Clearly, the high quality photographs gets higher scores in general, compared to the low quality photographs of each category photographs. Next we analyze the results of the experiments.
\begin{table}
\begin{center}
\caption{Accuracy of the proposed method across various categories}
\renewcommand{\arraystretch}{1.05}
\begin{tabular}{|m{2cm}|m{2cm}|}
\hline
\textbf{Category} & Accuracy\\ \hline
Animal & 89{\%}\\ \hline
Architecture & 84{\%}\\ \hline
Human & 93{\%}\\ \hline
Landscape & 83{\%}\\ \hline
Night & 76{\%}\\ \hline
Plant & 91{\%}\\ \hline
Static & 89{\%}\\ \hline
Overall & 87{\%} \\ \hline
\end{tabular}
\end{center}
\label{tab2}
\end{table}

\subsection{Results and Discussions}
We measure the accuracy of the proposed method by calculating the percentage of correct classifications of the test photographs of the dataset. Table II shows the accuracy of the proposed method on each category of photographs separately, alongwith the overall accuracy on the whole test set. The overall accuracy of the proposed method is increased by 6\% compared to \cite{tang}, the most recent effort made to work on this specific problem. All the accuracy measures shown in Table II, are provided corresponding to $\theta=4$.

\begin{figure*}
\centering{
\includegraphics[height=4cm, width=4.4cm]{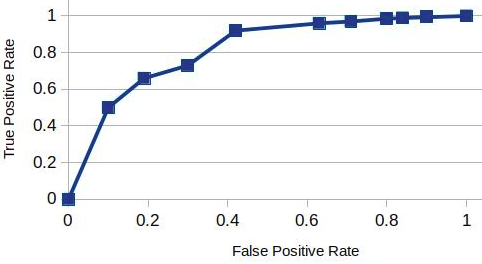}
\includegraphics[height=4cm, width=4.4cm]{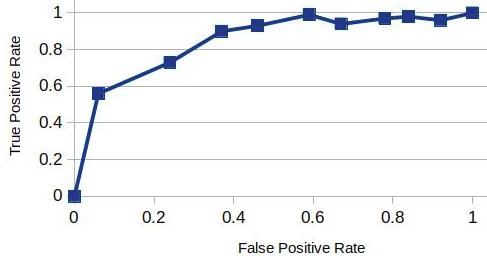}
\includegraphics[height=4cm, width=4.4cm]{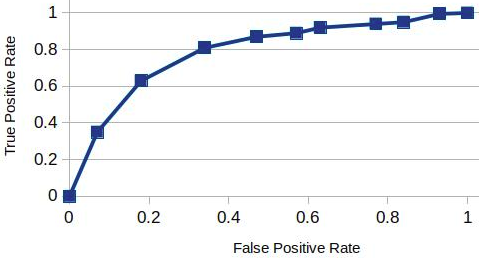}
\includegraphics[height=4cm, width=4.4cm]{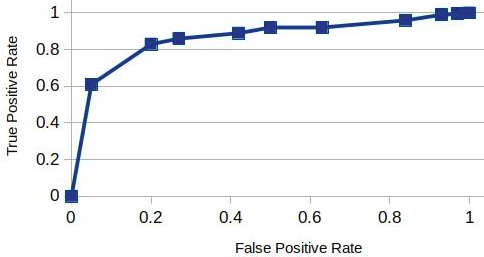}}
\centerline{(a)\hspace*{1.5in}(b)\hspace*{1.5in}(c)\hspace*{1.5in}(d)}
\centering{
\includegraphics[height=4cm, width=4.4cm]{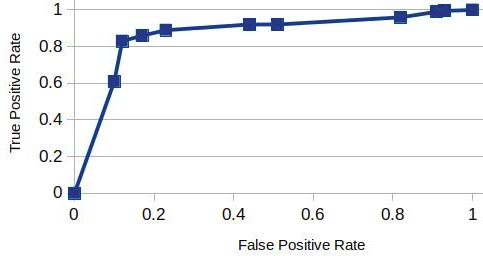}
\includegraphics[height=4cm, width=4.4cm]{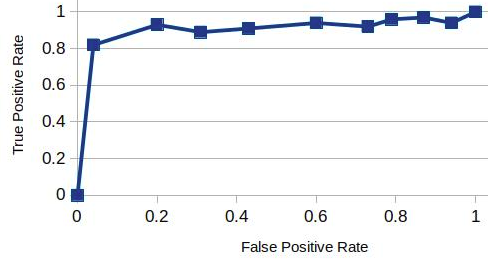}
\includegraphics[height=4cm, width=4.4cm]{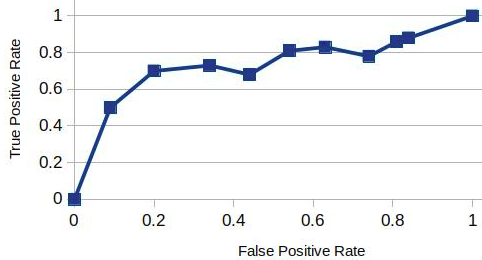}
}
\centerline{(e)\hspace*{1.75in}(f)\hspace*{1.75in}(g)}
\caption{ROC curves drawn on the performance of the proposed approach on each category separately: (a) Animal, (b) Landscape, (c) Architecture, (d) Plant, (e) Static, (f) Human and (g) Night photographs.}
\label{roc}
\end{figure*}

We show the ROC curves separately for each category of photographs, in Fiure \ref{roc}. The ROC curves are obtained by changing the discrimination constant. The area under the curve measured for the ROC curves drawn for different categories of photographs are given as follows: Animal 0.8138, Architecture 0.7865, Static 0.8504, Night 0.7298, Human 0.9013, Plant 0.8629 and Landscape 0.8472.

\begin{figure}
\centering{
\includegraphics[height=4cm, width=4.2cm]{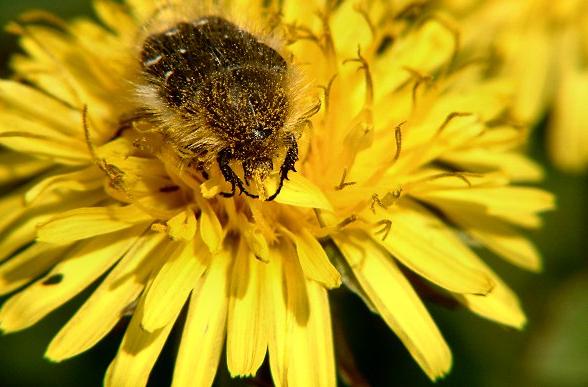}
\includegraphics[height=4cm, width=4.2cm]{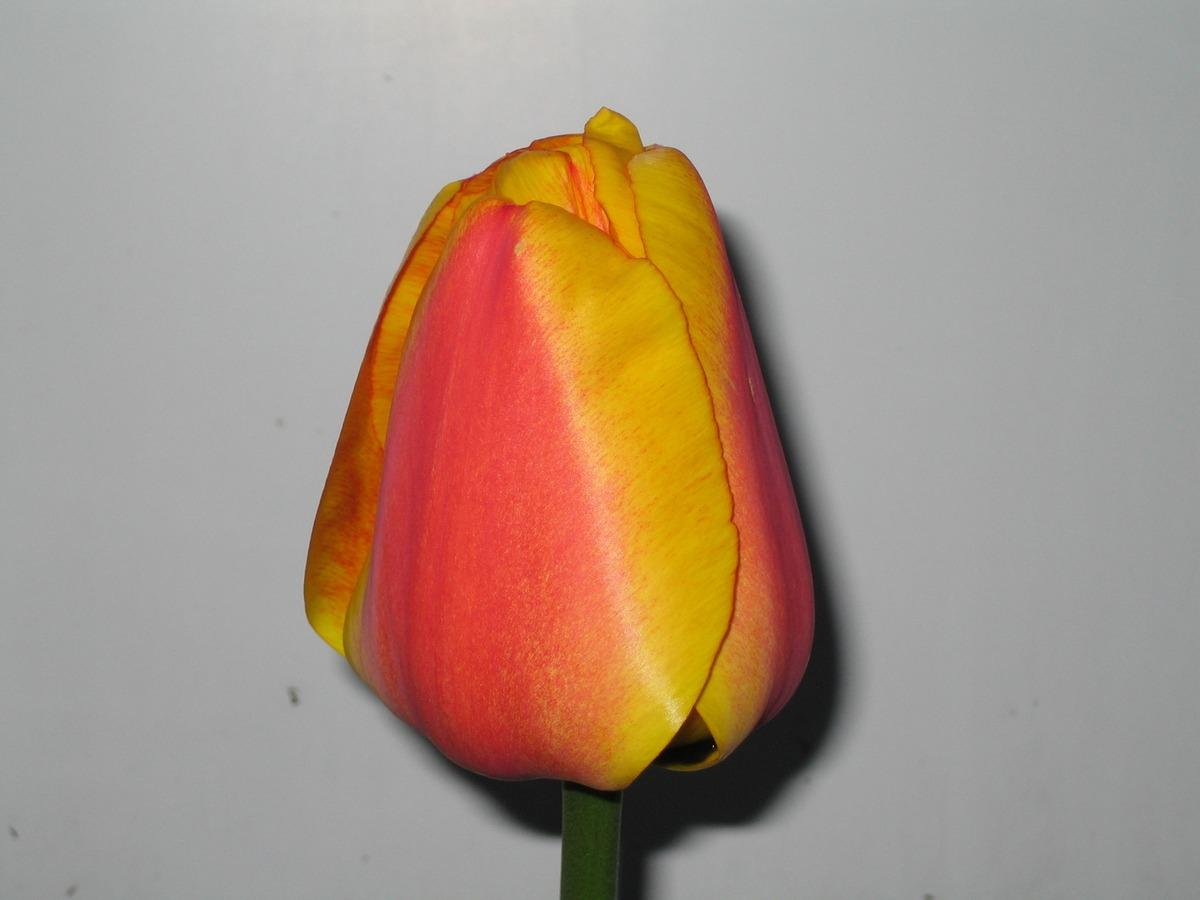}}
\centerline{(a)\hspace*{1.5in}(b)}
\caption{(a) A high quality image but quality score is 3.98 (b) A low quality image with MLP score 5.42.}
\label{fail}
\end{figure}

Figure \ref{fail} shows two examples of failure of the proposed method: the first one is an example of false positive (i.e., the photograph is captured by a professional photographer, but is detected as a low quality image by the proposed method) and the second photograph is an exaple of false negative (i.e., the photograph is captured by a common people, but is detected as a high quality image by the proposed method). In the first case, a portion of the object of interest (the flower) is blured, which may be the cause of getting a low quality score by the proposed method, even though the photo is captured by a professional. In the second case, the object of interest (the flower) is appeared with sharp edges and good color combination, which may be the cause of getting a high score by the proposed method.

\section{Conclusions and Future Scopes}
\label{conclude}
We have proposed a technique for assessment of the artistic quality of photographs, by introducing suitable weightage to some effective visual featues extracted from the image. The proposed method can provide a quality score for the photograph. It will be a hectic task to manually label the photographs based on the results extracted. To improve the accuracy and efficiency of the grading systema semi-supervised learning technique may be applied on Multi-layer perceptron in order to accomodate the bulk data which are unlabelled but they can contribute significantly in the learning. Another potential area of future research may be extending this two-class classification system to a multi-class classification system, to categorize different qualities of photographs, which may be used for automatic photo indexing.

\end{document}